\documentclass[conference]{IEEEtran}
\IEEEoverridecommandlockouts
\usepackage{cite}
\usepackage{amsmath,amssymb,amsfonts}
\usepackage{algorithmic}
\usepackage{graphicx}
\usepackage{textcomp}
\usepackage{xcolor}
\usepackage{algorithm,algorithmic,soul}
\def\BibTeX{{\rm B\kern-.05em{\sc i\kern-.025em b}\kern-.08em
    T\kern-.1667em\lower.7ex\hbox{E}\kern-.125emX}}
\newcommand{\Nbest}{\mathbf{N}_{\text{best}}}
\newcommand{\bx}{\mathbf{x}}
\newcommand{\bv}{\mathbf{v}}

\begin{document}

\title{Heterogeneous Swarms for Maritime Dynamic Target Search and Tracking\\
\thanks{This research is partially funded by Thales Solutions Asia Pte Ltd, under the Economic Development Board - Industrial Postgraduate Programme (EDB-IPP), and the SUTD SGP-AI Grant}
}

\author{\IEEEauthorblockN{Hian Lee Kwa}
\IEEEauthorblockA{\footnotesize Singapore University of Tech. and Design\\ 
Thales Solutions Asia Ptd Ltd\\
Singapore \\
hianlee\_kwa@mymail.sutd.edu.sg}
\and
\IEEEauthorblockN{Grgur Toki\'c}
\IEEEauthorblockA{\footnotesize Dept. of Mechanical Engineering\\
Massachusetts Institute of Technology\\
Cambridge, MA, USA\\
gtokic@mit.edu}
\\
\IEEEauthorblockN{Dick K. P. Yue}
\IEEEauthorblockA{\footnotesize Dept. of Mechanical Engineering\\
Massachusetts Institute of Technology\\
Cambridge, MA, USA\\
yue@mit.edu}
\and
\centering
\IEEEauthorblockN{Roland Bouffanais}
\IEEEauthorblockA{\footnotesize Engineering Product Development\\
Singapore University of Tech. and Design \\
Singapore \\
bouffanais@sutd.edu.sg}
}

\maketitle

\begin{abstract}
Current strategies employed for maritime target search and tracking are primarily based on the use of agents following a predetermined path to perform a systematic sweep of a search area. Recently, dynamic Particle Swarm Optimization (PSO) algorithms have been used together with swarming multi-robot systems (MRS), giving search and tracking solutions the added properties of robustness, scalability, and flexibility. Swarming MRS also give the end-user the opportunity to incrementally upgrade the robotic system, inevitably leading to the use of heterogeneous swarming MRS. However, such systems have not been well studied and incorporating upgraded agents into a swarm may result in degraded mission performances. In this paper, we propose a PSO-based strategy using a topological $k$-nearest neighbor graph with tunable exploration and exploitation dynamics with an adaptive repulsion parameter. This strategy is implemented within a simulated swarm of 50 agents with varying proportions of fast agents tracking a target represented by a fictitious binary function. Through these simulations, we are able to demonstrate an increase in the swarm's collective response level and target tracking performance by substituting in a proportion of fast buoys.
\end{abstract}


\section{Introduction}
The search and tracking of targets has many applications within the field of oceanic engineering. Examples of this include search and rescue~(SAR)~\cite{Lopez2017}, environmental monitoring~\cite{Coogle2013}, and underwater mine countermeasures~(UMCM)~\cite{Tan2004}. Recently, there has been increasing interest in the deployment of autonomous multi-robot systems~(MRS) to mitigate the risks presented to human operators in hazardous working environments, as well as to reduce the need for human intervention in large scale and routine operations. Currently, the leading strategy for maritime target search and tracking is a systematic sweep of a search area carried out by agents following a predetermined route~\cite{Meghjani2016, Sousselier2015}. However, these types of strategies are inadequate when dealing with targets that are able to travel faster than any individual agent. In addition, these strategies are not scalable and are unable to cope with the sudden failure of individual agents.

To overcome these critical issues, recent studies have focused on the implementation of decentralized multi-agent control strategies for target search and tracking using swarming MRS~\cite{Senanayake2016, Coquet2019, Kwa2020}. One of the most popular strategies in this field is based on Particle Swarm Optimization~(PSO)---originally a computational method that simulates the dynamics of discrete agents behaving like flocking animals for the optimization of nonlinear functions. To implement this strategy in robots, several modifications have been made to the original PSO algorithm, such as the implementation of obstacle avoidance~\cite{Couceiro2011} and taking into account the limited communications range of robots~\cite{Coquet2019, Yang2019}. MRS using decentralized strategies such as PSO present several advantages over their traditional path following counterparts, as well as other non-decentralized or non-swarming MRS. These include faster search times, the ability to tolerate the failure of one or multiple robots---robustness, the ability to carry out the assigned mission with differing number of robots---scalability, and the ability to perform a task in a dynamic environment---flexibility~\cite{Bouffanais2016, Kit2019}.

In addition, the modularity of MRS presents the opportunity for the incremental upgrade of the individual agents. Upgrades of this nature will inevitably result in a heterogeneous MRS comprised of agents with different physical and computational processing abilities. However, the incorporation of improved agents within a swarm may not necessarily benefit the swarm and could, conversely, be detrimental to the performance of the MRS~\cite{Altshuler2009}. For example, it was demonstrated that in the dynamic monitoring of open water bodies using mobile buoys, the addition of a small number of faster travelling buoys resulted in a significant increase in the swarm's collective coverage performance only when the faster moving buoys were placed in the vicinity of the slower moving buoys~\cite{Vallegra2018}. A previous study has also suggested that there may be an optimum composition proportion in which heterogeneous teams should be deployed when carrying out site defence missions~\cite{Strickland2018}. 

Furthermore, current research in heterogeneous PSO mainly focuses on different agent behaviors and communications networks instead of the motile properties of heterogeneous agents~\cite{Nepomuceno2013, Ma2016}. This is due to the original applications of PSO \textit{in silico}, where the agents' travelling speeds are irrelevant and therefore ignored. As such, it is necessary to study physically heterogeneous swarms in order to benefit from the different capabilities of each agent in the physical world.

Current literature on PSO-based robotics is also mostly concentrated on the search for static targets, leaving the implementation of these behaviors for dynamic target searches largely unexplored. The two main challenges associated with dynamic target search and tracking using PSO-based algorithms are: (1) the swarm's use of outdated information, and (2) a lack of exploratory actions carried out by the system~\cite{Jordehi2014}. In both cases, the swarm may be able to initially converge on the target but will be unable to continue tracking the target's movements once engaged due to the PSO algorithm's tendency to favor local exploitation over global exploration during the latter stages of the search process~\cite{Hussain2019}. This shows that the swarm needs to be able to constantly explore the search space even after the target is found. In PSO-based robotics, common methods that promote search space exploration are done through the implementation of a memory limit, in which outdated information is discarded~\cite{Coquet2019} or by employing inter-agent repulsion, preventing individuals from clustering in a small area~\cite{Blackwell2018}.

Recently, it has been uncovered that there exists an optimal level of connectivity between swarming agents involved in a decentralized decision-making process, regardless of the agents' physical capabilities~\cite{Mateo2019}. This optimal level of connectivity is directly related to the speed of the driving signal of the target. In addition, both excessive or insufficient levels of inter-agent connectivity was found to limit a swarm's ability to collectively respond to perturbations~\cite{Mateo2017}. This implies that while tracking a dynamic target, tuning the amount of interaction between a swarm's may result in an improved collective tracking performance.

In light of this, we recently proposed a PSO-based strategy using a topological $k$-nearest neighbor graph with tunable exploration and exploitation dynamics (EED) with an adaptive repulsion parameter~\cite{Kwa2020}. It has also been shown that by keeping the neighbors in communications range constant, the level of consensus reached by the swarm was able to be maintained despite sudden removals and injections of large amounts of agents~\cite{Rausch2019}. Therefore, by means of a topological $k$-nearest neighbor network, the level of inter-agent information exchange can be controlled better in comparison to a network determined purely by metric distance, maintaining the performance of the swarm~\cite{Komareji2017}.  In addition, by using the adaptive repulsion parameter, we are able to promote exploration of the search area among the swarming agents, preventing them from aggregating in a small area. Through this strategy, it was found that an optimum level of connectivity exists, at which the performance of a homogeneous swarm tracking a fast moving target was maximized.

In this paper, we present an extension of our work in~\cite{Kwa2020}, which used a homogeneous PSO-based swarm with an adjustable EED. Here, a heterogeneous swarm of 50 agents, comprised of two different types of agents with different maximum speeds, is modelled to search and track a target represented by a fictitious binary objective function. In doing so, we emulate one of the challenging cases of searching for a target travelling faster than the agents themselves while using a near-zero-range sensor. The performance of various swarm compositions is compared through the use of several metrics, providing insights into the swarm's tracking performance, its response to the moving target, as well as its EED performance.

\section{Methods}

\subsection{Search and Track Strategy}
To achieve our goal of searching for and tracking a mobile target, we have implemented a PSO-based swarming MRS with tunable EED. This solution relies on a decentralized PSO algorithm using a dynamic network, searching for the global minimum location that corresponds to the position of a target. In addition, an adaptive repulsion behavior that increases the amount of collective exploratory actions performed by the swarm is implemented. This last parameter is controlled by varying the degree $k$ of the swarm's $k$-nearest neighbor communications network. 

\subsubsection{$k$-Nearest Neighbor Particle Swarm Optimization}\label{sec:pso}
In particle swarm optimization, the $N$ agents (also commonly referred to as particles or candidate solutions) that comprise a swarm are initially randomly placed around the search space. At any discrete time-step $t$, each agent $i$ can be fully characterized by three state variables: its two-dimensional velocity, $\bv_i[t]$, its position, $\bx_i[t]$, and its objective function value, $f(\bx_i[t],t)$. In this paper, the objective function value of an agent $i$ at any time-step $t$ is assigned as follows. 

\begin{equation}
f(\bx_i[t],t)=\begin{cases}
-1 & \text{ if agent } i \text{ is on target}, \\
0 & \text{ otherwise}.
\end{cases}
\label{eqn:func_val_assign}
\end{equation}

Assigning an objective function value of either $-1$ or $0$, as seen in~\eqref{eqn:func_val_assign}, allows a binary objective function to be modelled. The explicit dependence on time of the objective function value, $f(\bx,t)$, reflects the dynamic character of the target.

At each time-step, each agent $i$ attempts to minimize its objective function value by taking into account its current direction of travel and the best position of an agent within its neighborhood (also known as the ``Neighborhood best''), denoted here by $\Nbest$. This variable, also known as the social component, is the main driver behind the exploitative actions carried out by the swarm. Classical variants of PSO also include the influence of the personal best position---known as the cognitive component, which is reliant on the memory of the agents and causes them to explore the search area more~\cite{Kennedy1995}. Given that one the most challenging cases of a dynamic objective function is considered, the effects stemming from the utilization of an agent's memory are known to be completely counter-effective and are therefore discarded as the information previously collected by the swarm rapidly becomes outdated~\cite{Leonard2011, Coquet2019}. The velocity and position of the agents are updated according to Eqs.~\eqref{eqn:vel_update}~\&~\eqref{eqn:pos_update}. The parameter $\omega$ is known as the velocity inertial weight, and $c$ as the velocity social weight, while $r$ is a number randomly drawn from the unit interval.
\begin{align}
    \bv_{\text{pso},i}[t+1] &= \omega \bv_i[t] + c r \big(\Nbest[t+1] - \bx_i[t+1]\big),
    \label{eqn:vel_update}\\
    \bx_i[t+1] &= \bx_i[t] + \bv_i[t]. 
    \label{eqn:pos_update}
\end{align}
It should also be noted that $\Nbest[t]$ is assigned the position of an agent's topological neighbor with the best position at any particular time-step. This is unlike the original PSO algorithm where $\Nbest[t]$ takes on the historical best position found by the particle's neighbors. It is important note that in our framework, the concept of a neighbor is understood in the network sense~\cite{Sekunda}. This means that any agent $i$ has as many neighbors as its degree $k_i$, and that the neighborhood changes given that a time-varying network topology is considered.

The idea of a topological neighborhood is conveniently represented by a $k$-nearest neighbor graph. Here, the value of the degree $k$, which is the same for all agents, characterizes the level of connectivity within the swarm. At each time-step, all agents identify their $k$-nearest topological neighbors, and set their $\Nbest$ position to be the position of the nearest neighbor on target. Should the agent detect the target itself, or should none of the agent's neighbors detect the target, $\Nbest$ is then set to the agent's own position $\bx_i$ at that given time-step. At this point, $\mathcal{N}_i[t]$ is defined as the set of $k$ indices corresponding to the topological neighbors of agent $i$ at time-step $t$. In the sequel, the reference to the time-step is dropped to simplify the notations.

It has previously shown that the level of connectivity (degree) significantly affects the collective dynamics or a swarm~\cite{Bouffanais2016}. In addition, it has also been found that for PSO in both dynamic and static environments, higher levels of connectivity lead to higher levels of local exploitation, and consequently a higher degree of aggregation within the search space, while a low degree causes the swarm to prioritize exploration of the environment~\cite{Blackwell2018, Kwa2020}.

\subsubsection{Adaptive Repulsion}
The adaptive repulsion behavior was introduced in~\cite{Kwa2020}. Through its implementation, the swarm was encouraged to carry out more exploratory movements, preventing agents from congregating within a small area after being outrun by their target. This led to an increase in the swarm's tracking performance of the dynamic target when compared to a swarm using a constant repulsion scheme and also resulted in an increase to the swarm's responsiveness to the changing environment. Furthermore, this behavior also provides a welcome inter-agent collision-avoidance measure.

The repulsive behavior adopted was based on the one used in another swarming system~\cite{Vallegra2018, Zoss2018}. For any agent $i$ with topological neighbors $j$, the repulsion velocity can be expressed as:
\begin{equation}
    \bv_{\text{rep},i}[t] = - \sum_{j\in \mathcal{N}_i}\left( \frac{a_R[t]}{r_{ij}[t]}\right)^d \frac{{\mathbf{r}_{ij}[t]}}{r_{ij}[t]}, 
    \label{eqn:rep}
\end{equation}
where $\mathbf{r}_{ij}$ is the vector from agent $i$ to agent $j$. This level of inter-agent repulsion is controlled by two variables: the repulsion strength $a_R$, affecting the agents' distance from each other at equilibrium, and the exponent $d$ in the pre-factor term $(a_R/r_{ij})$. In the simulations carried out, $d$ is fixed at $6$ given that this value has very moderate effects on the performance of the EED strategy. At large $(a_R/r_{ij})$ and $d$ values, the repulsion strength causes the agents to distribute themselves evenly across the environment~\cite{Vallegra2018}. As such, these two parameters were set in a manner that enabled the swarm to cover the entire search area when there was no target found.

Critical to the implementation of the repulsive behavior is the ability of an agent to vary its level of repulsion based on the information it collects from the environment and provided by its neighbors. From these two sets of information, each agent can adjust the value of its own parameter $a_R$ within a set range, thereby yielding adaptive repulsion. In our numerical tests, $a_R$ was allowed to vary between 0.375 and 1.5. If the target is not within detection range of an agent or its neighbors, the agent enters an exploratory state, gradually increasing its $a_R$ until the maximum limit for $a_R$ is attained. On the other hand, if an agent or at least one of its neighbors is able to detect the target, the agent adopts a tracking state where the strength of its repulsion is lowered until a set minimum value. The adaptive repulsion scheme is summarized in Algorithm~\ref{adaptive_repulsion}.

\begin{algorithm}
    \caption{Adaptive Repulsion}
    \label{adaptive_repulsion}
    \begin{algorithmic}
    \STATE Set $a_{R,\text{min}} = 0.375$, $a_{R,\text{max}} = 1.5$, $d=6$, and $\delta=0.01$
    \WHILE{System active}
    \IF{$f(\bx_i[t], t) = -1$ \OR $\exists j \in \mathcal{N}_i$ s.t. $f(\bx_j[t], t) = -1$}
        \IF{$a_R > a_{R, \text{min}}$}
            \STATE $a_R \gets a_R - \delta$
        \ENDIF
    \ELSE
        \IF{$a_R< a_{R, \text{max}}$}
            \STATE $a_R \gets a_R + \delta$
        \ENDIF
    \ENDIF
    \STATE Calculate $\bv_{\text{rep},i}$ using \eqref{eqn:rep}
    \ENDWHILE
    \end{algorithmic}
\end{algorithm}

\subsubsection{Exploration Exploitation Dynamics}
Given the elements reported in the previous sections, the proposed search and tracking strategy is presented in Algorithm~\ref{strategy}.

\begin{algorithm}
    \caption{Dynamic $k$-Nearest Network PSO with Adaptive Repulsion}
    \label{strategy}
    \begin{algorithmic}
    \STATE Set $t = 0$, $k \in [2, N-1]$, $\omega=1$, and $ c=0.5$
    \WHILE{System active}
        \FOR{All agents $i \in [1, N]$}
            \STATE Calculate $f(\bx_i[t], t)$
            \IF{$f(\bx_i[t], t) = -1$}
                \STATE $\Nbest \gets \bx_i[t]$
            \ELSE
                \STATE Determine $\mathcal{N}_i = \{j \in [1, N]$ s.t. agent $j$ is a topological \textit{k}-nearest neighbor of agent $i$\}
                \IF{$\exists j \in \mathcal{N}_i$ s.t. $f(\bx_j[t], t) = -1$}
                    \STATE $\Nbest \gets \bx_j[t]$
                \ELSE
                    \STATE $\Nbest \gets \bx_i[t]$
                \ENDIF
            \ENDIF
            \STATE Calculate $\bv_{\text{pso},i}$ using \eqref{eqn:vel_update}
            \STATE Calculate $\bv_{\text{rep},i}$ using Algorithm~\ref{adaptive_repulsion}
            \STATE $\bv_i[t] \gets \bv_{\text{pso},i}[t] + \bv_{\text{rep},i}[t]$
            \STATE $\bv_i[t] \gets  \bv_i[t]/v_{\text{max}}$
            \STATE $\bx_i[t+1] \gets \bx_i[t] + \bv_i[t]$
        \ENDFOR
        \STATE $t \gets t+1$
    \ENDWHILE
    \end{algorithmic}
\end{algorithm}

\subsection{Swarm Performance Metrics}
To analyze the performance and dynamics of the swarm, three different metrics previously introduced in~\cite{Kwa2020} for homogeneous swarms are adapted and used. These are the cumulative velocity fluctuation magnitude, the correlation between an agent's heading and its bearing relative to the target's location, and the proportion of time the swarm has at least one agent in detection range of the target. 

\subsubsection{Cumulative Velocity Fluctuation Magnitude}\label{cfm}
It has previously been shown that in various flocking animals, including midges and starlings, the effectiveness of the transmission of information between swarming agents is closely connected to the level of velocity fluctuations of individual agents and their correlations~\cite{Cavagna2010, attanasi2014collective}. As such, the level of velocity fluctuations in an artificial swarm can be quantified to provide insight into the level of response of a swarm to its dynamic operating environment.

The velocity fluctuation of an agent \textit{i} is defined as the difference between its velocity and the mean velocity of the entire swarm:
\begin{equation}
\mathbf{u}_i[t] =\bv_i[t] - \langle \bv_j[t] \rangle_{j=1,\dots,N} =\bv_i[t] - \frac{1}{N} \sum_{j=1}^N \langle \bv_j[t] \rangle. 
\label{eqn:vel_diff}
\end{equation}
The average maximum speed of the swarm's comprising agents can also be calculated:
\begin{equation}
    \bar{v}_{\text{max}}=\frac{N_f v_{f,\text{max}}+N_s v_{s,\text{max}}}{N}.
    \label{eqn:avg_max_speed}
\end{equation}
Where $N_f$ is the number of fast agents, $v_{f,\text{max}}$ is the maximum velocity of the fast agents, $N_s$ is the number of slow agents, and $v_{s,\text{max}}$ is the maximum velocity of the slow agents.

The overall response of the swarm to its dynamic environment can then be determined by taking the time-averaged sum of all of the swarming agents' velocity fluctuations, divided by the number of agents in the swarm, and then normalizing the value with respect to the average maximum speed of the agents in the swarm:
\begin{equation}
\Xi = \frac{1}{T_f}\sum_{t=1}^{T_f} \left\langle \frac{\mathbf{u}_j[t]}{\bar{v}_{\text{max}}} \right\rangle_{j=1,\dots,N} =\frac{1}{NT_f\bar{v}_{\text{max}}} \sum_{t=1}^{T_f}\sum_{j=1}^N \mathbf{u}_j[t], \label{eqn:response}
\end{equation}
where $T_f$ is the total number of time-steps considered.

\subsubsection{Heading-Bearing Correlation}\label{hbc}
\begin{figure}[htbp]
    \centering
    \includegraphics[width=0.45\textwidth]{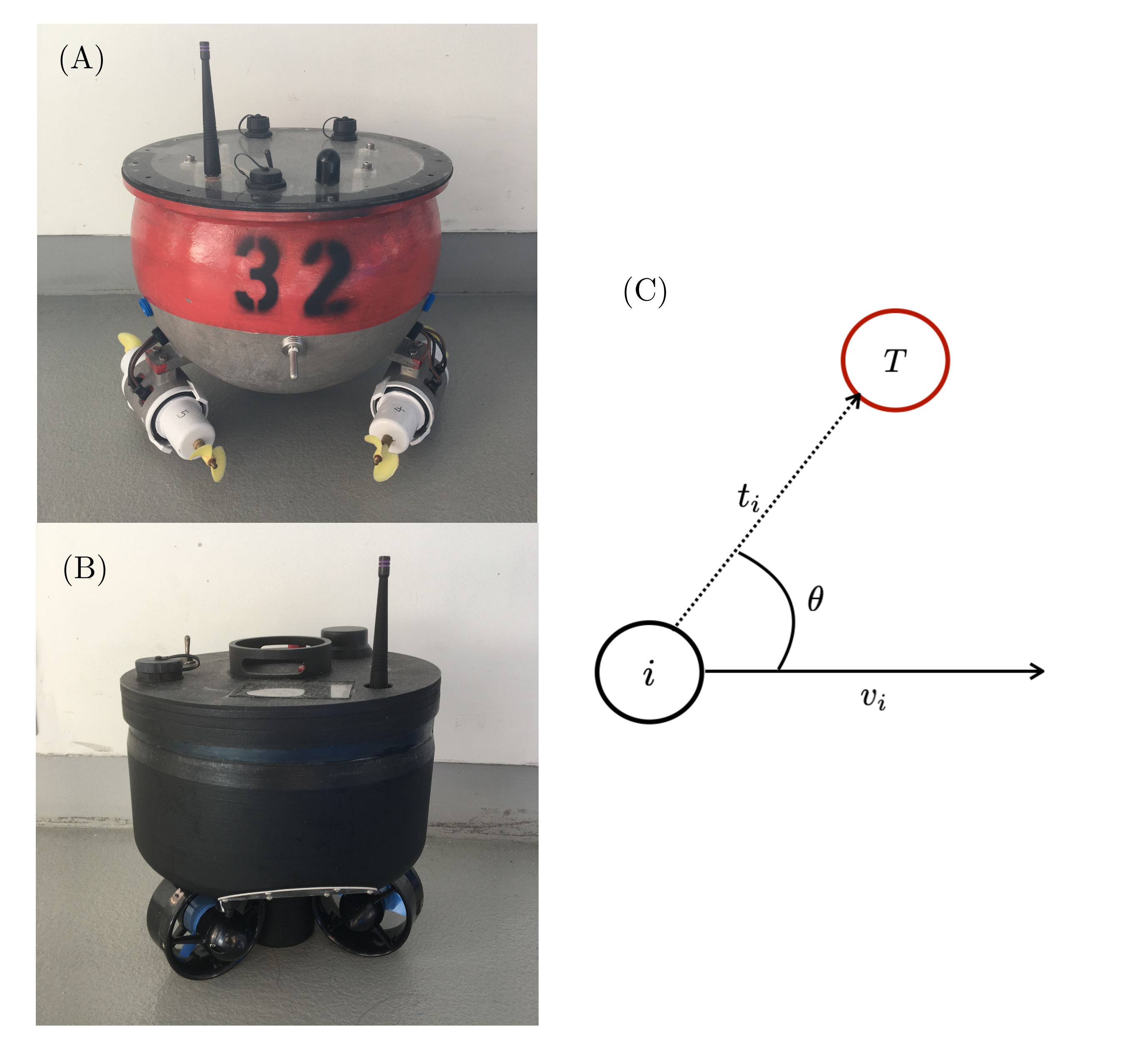}
    \caption{(A) Original BoB-0 unit. (B) Upgraded BoB-1 unit. (C) Bearing of the target ($T[t]$) relative to the buoy ($\mathbf{t}_i[t]$) and the heading of the buoy ($\bv_i[t]$) at time $t$.}
    \label{fig:buoys}
\end{figure}

To quantify the exploitative activity of the swarm, the correlation between the heading of any agent $i$ and the target's bearing relative to this agent at any time-step $t$, denoted by $\phi_i[t]$, is calculated. This correlation can be calculated as follows:
\begin{equation}
\phi_i[t] = \frac{\bv_i[t] \cdot \mathbf{t}_i[t]}{\left\|\bv_i[t]\right\|} = \cos \theta_i[t], \label{eqn:hbc}
\end{equation}
where $\mathbf{t}_i[t]$ is the bearing of the target in relation to the agent and $\theta_i[t]$ is the angle between the two unit vectors (see Fig.~\ref{fig:buoys}). The quantity $\phi_i$ will therefore take a value between $-1$ and $1$, with values approaching $-1$ being related to exploratory actions, and values approaching $1$ revealing exploitative actions carried out by agents moving towards the target. Calculating this variable for all agents at each time-step of the simulation allows the generation of a histogram, showing the overall EED of the entire swarm. It should be noted that with this formula, agents that are stationary will have a heading-bearing correlation value of $0$.

\subsubsection{Time on Target}\label{tot}
The overall tracking performance of the swarm is classically determined by counting the number of time-steps when the target is able to be detected by at least one agent. Using this metric enabled us to calculate the percentage of time the system is engaged in tracking the target as it moves around the search-space. 

\subsection{Swarm Robotic Platform}

In this work, we model the ``Bunch of Buoys'' (BoB) system, originally intended for the dynamic monitoring of open water bodies~\cite{Zoss2018, Vallegra2018}. The current system employs two types of buoys: BoB-0, which weighs $7.4~\text{kg}$ and can attain a maximum speed of $1.0~\text{m/s}$, and BoB-1, weighing $3.2~\text{kg}$ and is capable of attaining a top speed of $2.6~\text{m/s}$. Being faster and lighter, BoB-1 is able to better react to dynamic changes in its operating environment compared to BoB-0. Both of the robotic platforms have propulsion systems allowing for omnidirectional movement. In addition, both platforms house a suite of sensors, enabling the characterization of their local environment. A distributed mesh communications system is used for sharing locally sensed information among the buoys, as well as for sending and receiving of commands. In the simulations, 50 buoys tracking a fictitious moving target were modelled. Swarms with various compositions of BoB-0 and BoB-1 units and levels of connectivity were tested to observe the effects of gradually adding upgraded units to the system.

\subsection{Target Representation}
Most research on collective search and tracking considers cases involving a target emitting a continuous gradient field. In this case, a classical gradient-descent approach can be used to locate the position of the target with relative ease without utilizing swarm intelligence. To demonstrate the use of swarm intelligence, the target was represented through the use of a binary objective function. As explained previously in Section~\ref{sec:pso}, an agent on target was assigned an objective function value of $-1$ and $0$ otherwise. In the simulations, the target was also modelled to have maximum speeds greater than that of the maximum speed of the BoB-1 unit. Allowing the target to move at such speeds and using a binary objective function was meant to mimic one of the most challenging cases with a near-zero-range sensor tracking a target faster than the agent themselves. We believe that the design of effective swarm strategies benefits from considering such challenging scenarios.

\section{Simulation Results}
\subsection{Simulation Setup}
\begin{figure}[htbp]
    \centering
    \includegraphics[width=0.5\textwidth]{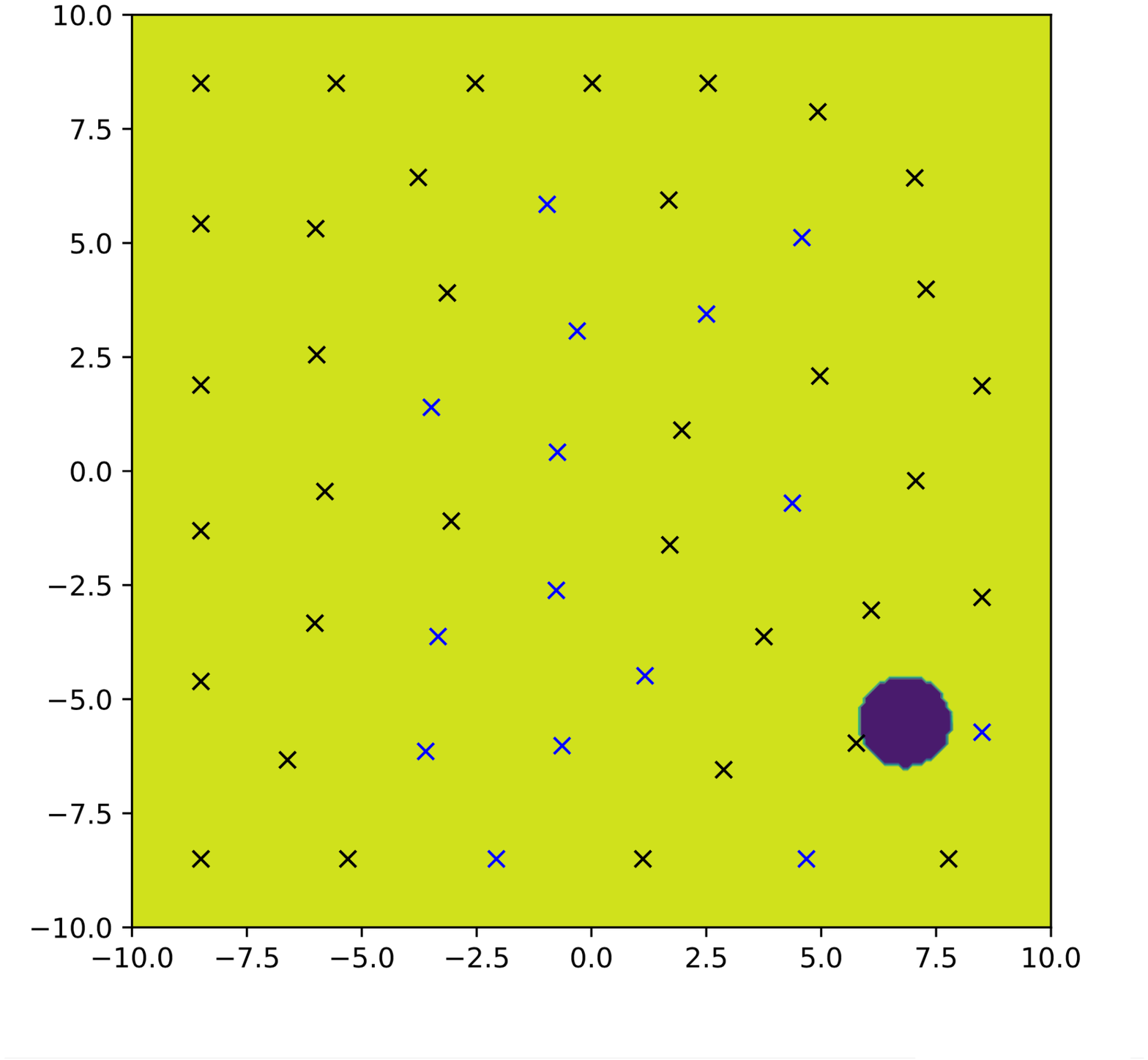}
    \caption{A heterogeneous swarm comprised of 35 BoB-0 units (black crosses) and 15 BoB-1 units (blue crosses) tracking a disc-shaped target (dark blue). Agents to the left of the target are locked in position due to them attaining their maximum repulsion strength and lack of target information.}
    \label{fig:locked}
\end{figure}

A two-dimensional square search-space (dimensions $L\times L$) was considered with a disc-shaped target having a fixed radius $\rho=L/20$ (see Fig.~\ref{fig:locked}). The target moved randomly around the search-space at constant speed $v$. If an agent fell within the radius of the target, it was assigned an objective function value of $-1$, and zero otherwise. This modelled a binary objective function, in which agents either were either fully informed of the target's position or had no information at all. In our simulations, the speed of the target was set to either $3.0 \text{ m/s}$ and $5.0 \text{ m/s}$, compared to $1.0 \text{ m/s}$ for the BoB-0 units and $2.6 \text{ m/s}$ for the BoB-1 units. All simulation runs lasted for a total of $100,000$ time-steps to ensure achieving statistically steady conditions. By using long simulation runs, a low level of variability was calculated for the three performance metrics between five different runs (below $0.3\%$ for both the tracking performance and the cumulative velocity fluctuation magnitude).

\subsection{Heading-Bearing Correlation}
\label{sec:eed}
\begin{figure}[htbp]
    \centering
    \includegraphics[clip,width=0.45\textwidth]{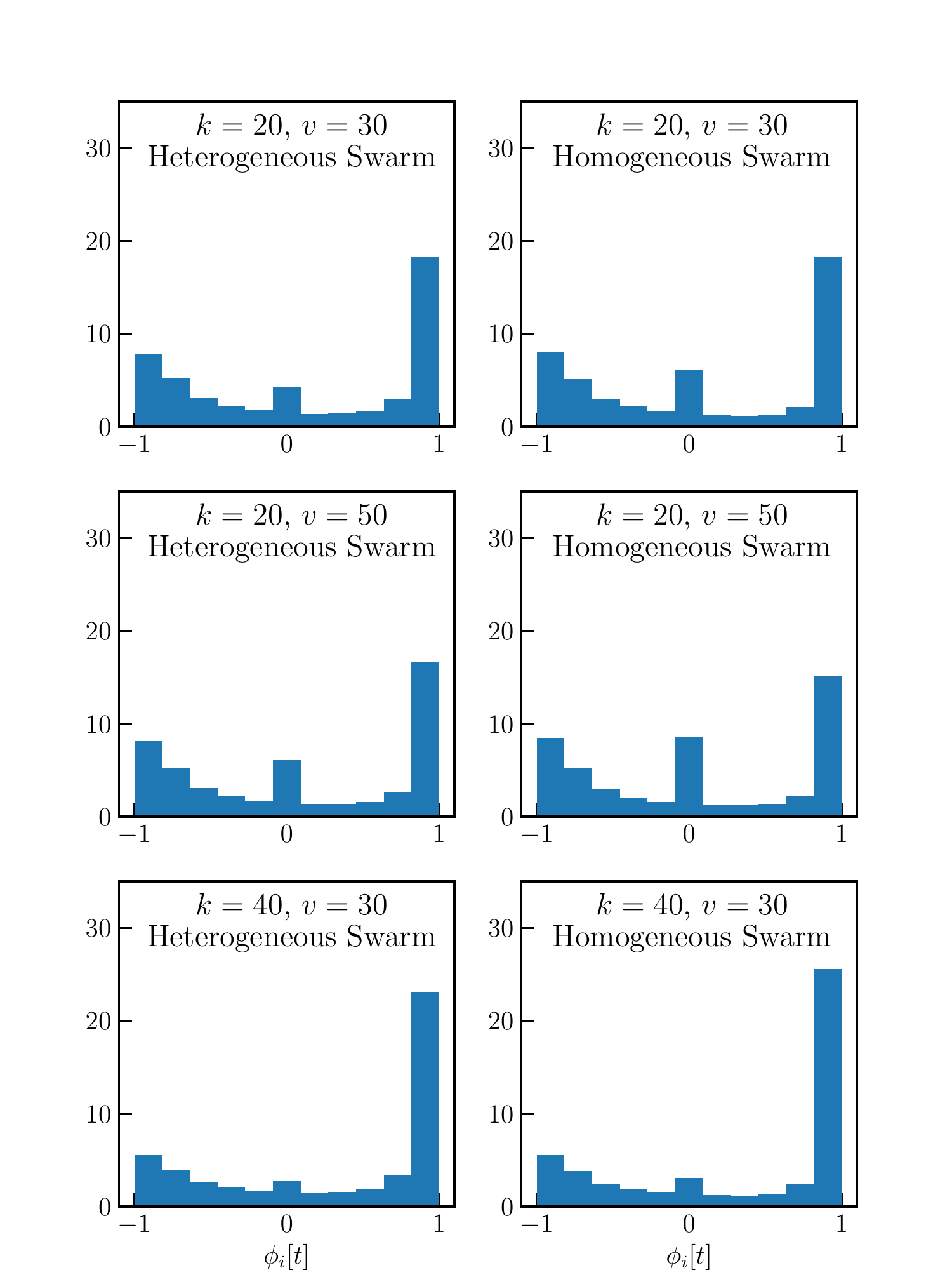}
    \caption{Distribution of heading-bearing correlations $\phi_i [t]$ ($x$-axis) for different levels of the swarm connectivity ($k$) and target speeds ($v$). The left column shows the heading-bearing correlations for a heterogeneous swarm comprised of 15 BoB-1 units and 35 BoB-0 units, while the right column shows the heading-bearing correlations for a homogeneous swarm of 50 BoB-0 units.}
    \label{fig:histogram}
\end{figure}

To visualize the overall EED of the swarm, the distribution of the heading-bearing correlation $\phi_i [t]$ for all agents $i$ in the swarm was considered over the entire duration of a simulation ($100,000$ time-steps). This was done by recording the heading-bearing correlations, calculated using Eq.~\eqref{eqn:hbc}, for all agents and iterations. The weight of each bin was then divided by the total number of iterations to give a time-averaged histogram for the entire swarm (see Fig.~\ref{fig:histogram}).

From the histograms, three observations can be made. Firstly, as the level of connectivity increases, so too does the amount of exploitative behavior carried out by the swarm. This observation is true for both the heterogeneous and homogeneous swarms. The increase in exploitation is characterized by a higher proportion of agents having a value large positive heading-bearing correlation value, suggesting that there was a higher proportion of agents moving towards the target instead of exploring the environment.

Secondly, in both homogeneous and heterogeneous swarms, the average number of agents with a heading-bearing correlation of $0$ increases at higher target speeds, implying a greater number of stationary agents. At higher speeds, the target is able to evade the agents more easily, causing the agents to spread out across the search area and be ``locked'' in place when their repulsion strength reaches the maximum limit. The agents remain in this position until they receive new information of the target's position (see Fig.~\ref{fig:locked}).

Finally, it can be seen that at lower levels of connectivity ($k=20$), the time averaged number of agents with a heading-bearing correlation of $0$ was lower for the homogeneous swarm than that of the heterogeneous swarms. This can be attributed to higher collective response of the heterogeneous swarm at lower levels of connectivity compared to the homogeneous swarm. The difference in the the levels of collective response is further explored in Section~\ref{sec:cfm_d}.

\subsection{Time on Target}
\begin{figure}[htbp]
    \centering
    \includegraphics[width=0.5\textwidth]{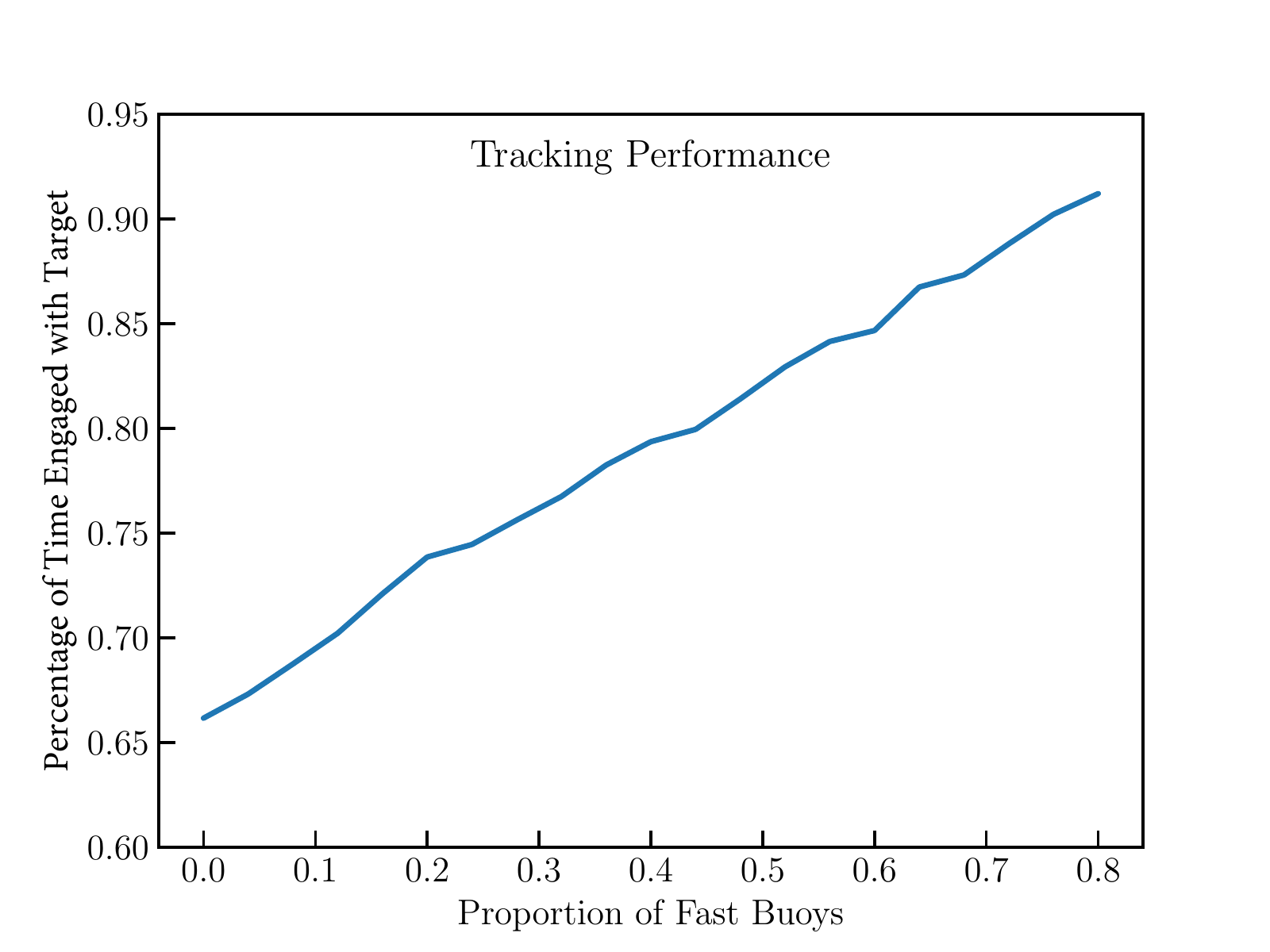}
    \caption{Tracking performance of a swarm comprised of varying proportions of fast buoys with a degree of $k=20$.}
    \label{fig:het_tot}
\end{figure}
\begin{figure}[htbp]
    \centering
    \vspace{-6ex}
    \includegraphics[width=0.5\textwidth]{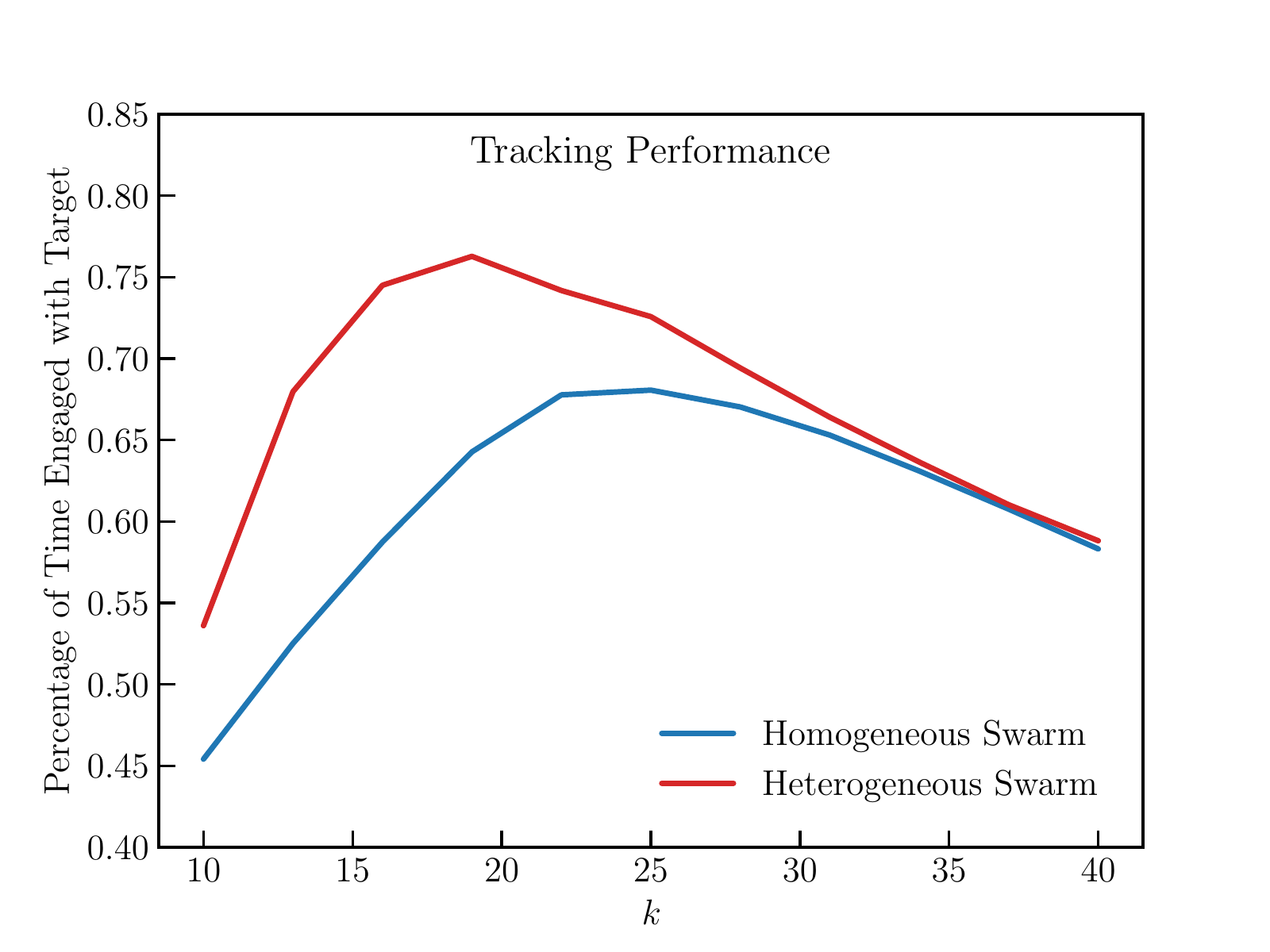}
    \caption{Tracking performance of a swarm of $50$ buoys comprised of $15$ fast agents with varying levels of connectivity, $k$.}
    \label{fig:tot_sweep}
\end{figure}
\begin{figure}[htbp]
    \centering
        \vspace{-6ex}
    \includegraphics[width=0.5\textwidth]{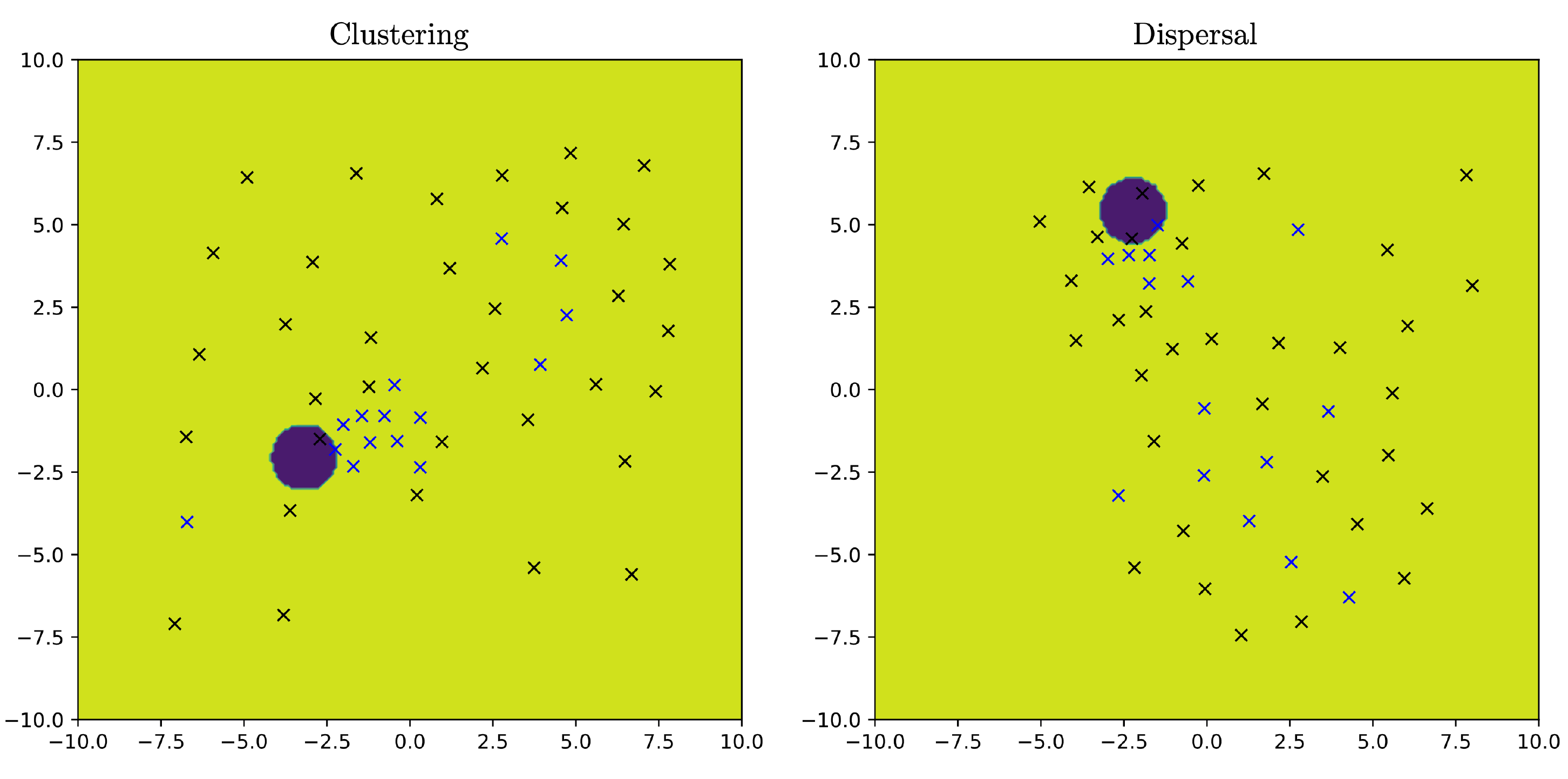}
    \caption{A heterogeneous swarm comprised of 35 BoB-0 units (black crosses) and 15 BoB-1 units (blue crosses) tracking a disc-shaped target (dark blue). (Left) BoB-1 units clustering together; (Right) BoB-1 units distributed throughout the swarm due to the implemented adaptive repulsion behavior.}
    \label{fig:clustering}
\end{figure}

The plots in Fig.~\ref{fig:het_tot} show that the overall tracking performance of the swarm increases with the proportion of fast buoys included within the swarm and is consistent across all levels of connectivity. This result is due to the faster buoys being better able to keep up with and respond to the presence of the fast moving target. However, the benefit of adding upgraded units decreases at higher levels of connectivity. Figure~\ref{fig:tot_sweep} also confirms that, similar to the homogeneous swarm, there exists an optimum degree of connectivity, $k^*\sim 18$, at which the tracking performance is at a maximum for the heterogeneous swarm. Beyond this level, higher amounts of connectivity are detrimental to the swarm's overall tracking performance due to the tendency of the swarm to carry out more exploitative actions, as mentioned in the previous section. Interestingly, this optimum occurs at a lower level of connectivity compared to the one for the slower homogeneous swarm. 

In the conducted simulations, despite the randomized starting positions of the buoys, the heterogeneous swarm always showed an improvement in performance over the homogeneous swarm. This implies that, unlike in dynamic area monitoring, the initial placement of the upgraded agents does not play an integral part in improving the swarm's tracking abilities. Through the visualization of the buoys' positions, as seen in Fig.~\ref{fig:clustering}, it can be observed that although the faster BoB-1 units may at times cluster together, they are eventually able to disperse themselves throughout the swarm. This is due to the implementation of the adaptive repulsion behavior that causes the individual agents within the swarm to have different levels of repulsion at any given time-step. As demonstrated in~\cite{Zoss2018}, depending on an agent's strength of repulsion with respect to its physical neighbours, the agents with different repulsion strengths can move past their neighbors by circumventing them or ``sneaking'' between them, instead of being locked within a fixed relative position.

\subsection{Cumulative Velocity Fluctuation Magnitude}
\label{sec:cfm_d}
\begin{figure}[htbp]
    \centering
    \vspace{-3ex}
    \includegraphics[width=0.5\textwidth]{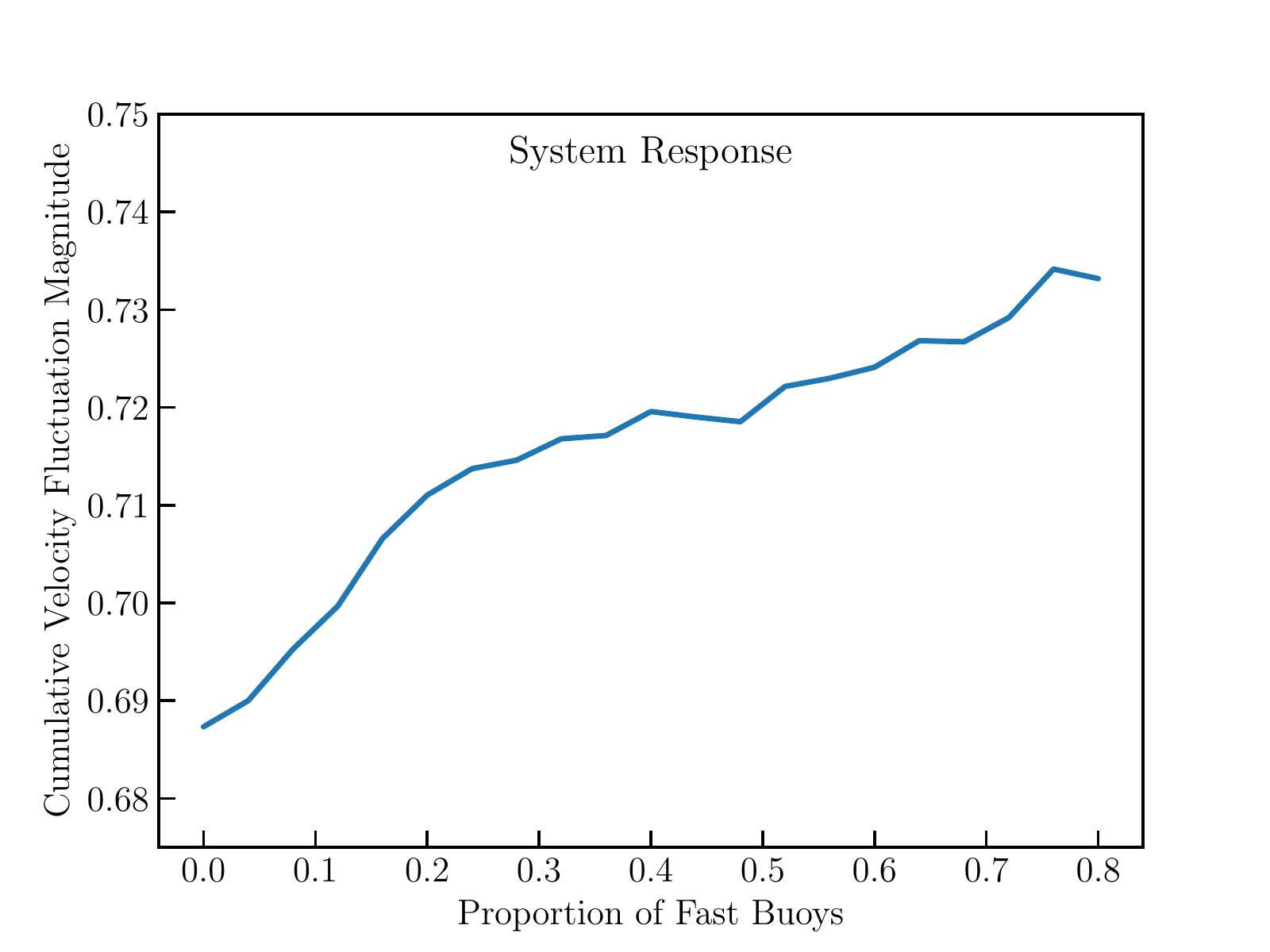}
    \caption{Cumulative velocity fluctuation magnitude plot of a swarm of varying proportions of fast buoys with a degree of $k=20$.} 
    \label{fig:het_cfm}
\end{figure}
\begin{figure}[htbp]
    \centering
    \includegraphics[width=0.5\textwidth]{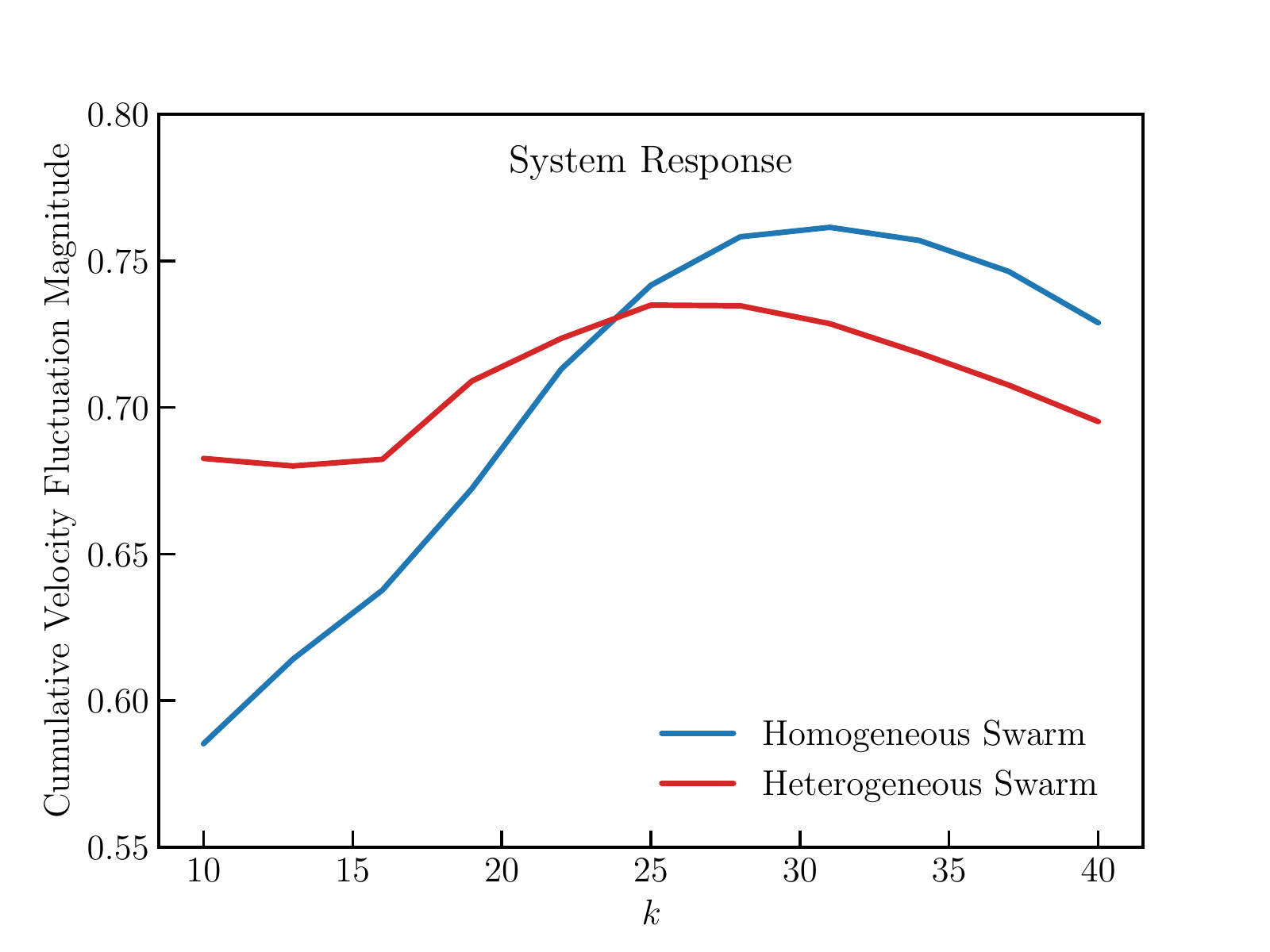}
    \caption{Cumulative velocity fluctuation magnitude plot of a swarm of $50$ buoys comprised of $15$ fast agents with varying levels of connectivity, $k$.} 
    \label{fig:cfm_sweep}
\end{figure}

As previously explained, the cumulative velocity fluctuation magnitude can be used as a proxy to quantify the level of responsiveness of a swarm to its dynamic environment, and hence the level of swarm intelligence activity. From Fig.~\ref{fig:het_cfm}, it can be seen that from low to intermediate levels of connectivity, the level of responsiveness of the swarm gradually increased with an increasing proportion of faster agents. Similar to the swarm's tracking performance, this result stems from the upgraded agents' ability to travel faster. Hence, the upgraded buoys were better able to react to the presence or absence of the target and carry out the necessary exploratory or exploitative actions.

In addition, it can be seen that in both homogeneous and heterogeneous swarms (Fig.~\ref{fig:cfm_sweep}), there is a peak in the systems' levels of response at intermediate levels of connectivity. This peak occurs at a lower degree in heterogeneous swarms ($k \sim 25$) than in the homogeneous one ($k \sim 30$). This is because upgrading a portion of the swarming MRS with faster agents essentially increases the average speed of the swarm. Therefore, even though the environment evolves at the same speed, this evolution occurs at a different time-scale relative to the swarm. By adding faster agents, the swarm needs to readjust its communications channel to adequately and optimally respond to its environment.

It should also be noted that in both homogeneous and heterogeneous swarms, the optimum level of connectivity associated with maximum response is higher than the level of connectivity required for maximum tracking performance. When the swarming agents are connected to provide maximum collective response, a large proportion of the swarm is attracted to the target once it is found. Consequently, there are little to no agents remaining to perform area exploration. As such, the target is eventually able to outrun the tracking agents. Conversely, when the swarm's communications network is connected to provide maximum tracking performance, there are still agents free to perform exploratory actions. Hence, when the initial group of pursuing agents are outrun, the remaining agents are able to continue tracking the target.

\section{Discussion}

The implementation of a swarming MRS in a target search and tracking mission allows for improving the performance of the system, as well as guaranteeing the robustness, scalability, and flexibility. In addition, the modularity of swarming MRS presents the opportunity to incrementally upgrade individual agents, necessitating the use of heterogeneous swarms. In this work, we presented a PSO-based heterogeneous swarm with an adjustable EED, capable of searching for and tracking a target moving faster than any of the individual swarming agents. The EED of the swarm was tuned through the use of an adaptive repulsion behavior along with a PSO-based algorithm using a topological $k$-nearest neighbor network. These behaviors were implemented in simulated models of the BoB swarming buoy system.

The performance and swarm dynamics of the system were calculated through the use of three metrics, namely: the heading-bearing correlation, cumulative velocity fluctuation magnitude, and the percentage of time on target. Through simulations, these metrics provided insights into the swarm's EED, response to the swarm's dynamic environment, and the proportion of time where at least one agent is engaged with the target.

Using these metrics, it was observed that the addition of faster buoys within the swarm did not affect the EED of the swarm at all levels of connectivity; increasing the degree $k$ for both homogeneous and heterogeneous swarms increased the amount of exploitative actions carried out by the swarming agents while a decrease in $k$ resulted in the two swarms favoring area exploitation. The simulations performed were also showed an improvement in the tracking performance of the swarm, regardless of the degree of the swarm's interconnecting $k$-nearest neighbour network and starting position of the buoys. However, this benefit reduced at higher levels of inter-agent connectivity. In addition, it was revealed that there exists an optimum degree $k^*$ at which the swarm's tracking performance was at a maximum for both swarms. This optimum degree was found to be lower in heterogeneous swarms than in homogeneous swarms due to the higher levels of system response generated by the heterogeneous swarms at low to intermediate levels of connectivity. 

\vspace{12pt}

\end{document}